# Benchmarking of LLM Detection: Comparing Two Competing Approaches

**Thorsten Pröhl**
Technische Universität Berlin
t.proehl@tu-berlin.de

**Erik Putzier**
Technische Universität Berlin
erik.putzier@campus.tu-berlin.de

**Rüdiger Zarnekow**
Technische Universität Berlin
ruediger.zarnekow@tu-berlin.de

## Introduction

Transformer-based AI systems have made great progress in areas such as text processing and comprehension. These deep learning models enable text generation and form the basis of modern language models. The rapid development in recent years has led to large language models such as GPT-4, Gemini or VICUNA-13B. These LLMs can be used for various tasks, such as chatbots on websites or as QA systems.

In addition, text summaries or completely new texts can be created on any topic. Possible uses for these texts include SEO texts for websites, product reviews or tests, news articles, theses, or email replies. It is immediately apparent that the numerous advantages and possibilities make work easier and faster for many stakeholders. However, negative effects such as SEO spam, fake product reviews, spam and disinformation in news articles, questions about personal work in theses, or email phishing are also conceivable and observable. This leads to the need to determine whether a text was written by a human or a machine.

This article gives an overview of the field of LLM text recognition. Different approaches and implemented detectors for the recognition of LLM-generated text are presented. In addition to discussing the implementations, the article focuses on benchmarking the detectors. Although there are numerous software products for the recognition of LLM-generated text, with a focus on ChatGPT-like LLMs, the quality of the recognition (recognition rate) is not clear. Furthermore, while it can be seen that scientific contributions presenting their novel approaches strive for some kind of comparison with other approaches, the construction and independence of the evaluation dataset is often not comprehensible. As a result, discrepancies in the performance evaluation of LLM detectors are often visible due to the different benchmarking datasets. This article describes the creation of an evaluation dataset and uses this dataset to investigate the different detectors. The selected detectors are benchmarked against each other.

## Detection of LLMs

LLM-generated texts play an increasingly significant role in a variety of domains. The emergence of LLMs such as ChatGPT has ensured that LLM-generated textual content can be easily generated and found on a massive scale on social networks, websites, in academic or creative writing. This can lead to problems, especially when AI is used for political misinformation, impersonation and forgery (Pröhl et al. 2024). Detecting such texts plays a crucial role in reducing the potential damage caused by AI-generated texts. Therefore, several detectors of such texts have been implemented using many different approaches. This leads to the first research question (RQ 1): What are the current approaches for the detection of LLM-generated texts?

Biderman and Raff (2022) investigate the origin of the texts, while Zellers et al. (2019) deal with the investigation of spam, fake, and disinformation campaigns. Furthermore, the study of training data is important to find indicators that signal whether a text was written by a human or generated by an LLM (Bender et al. 2021; Crothers et al. 2022; Mirsky et al. 2023).

Text generated by earlier language models is relatively easy to identify, as these texts do not yet sound too natural. The development of transformer-based language models has greatly improved the generation of human-sounding text (Radford et al. 2019). This renders the original detection methods useless.

Current approaches can be divided into white box and black box methods. White box methods use knowledge of how the model creates sequences to generate a signal. Krishna et al. (2023) follow the recording approach and Kirchenbauer et al. (2023) show possibilities of watermarking during generation. However, both approaches require full control over the generative language model and therefore cannot be applied to ChatGPT texts from OpenAI, for example.

Black box methods focus on post-hoc detection of LLM-generated texts. This has the advantage that no collaboration with the original developers of the models is necessary, as is the case with watermarking, for example.

Black box detection methods differ in their underlying approach but can be divided into two categories. The first category involves trained models for detection. A pretrained language model is used and fine-tuned to perform a binary classification, where the two classes are LLM or non LLM (Solaiman et al. 2019; Yu et al. 2023; Zellers et al. 2019). Hu et al. (2023) describe a technique for adversarial training. Tian et al. (2023) investigate an abstention learning method. The contributions describe that, it is both possible to train the entire backbone and to use only a classifier trained on the previously frozen learned features in order to include API returns. (Verma et al. 2023).

The second category includes recognition approaches that consider statistical features and patterns of LLM-generated text. A key advantage is the training effort. Here, no or only very little training data is required, in this context Pu et al. (2022) note that the adaptation of newer models is better, faster and easier. Tian (2023), Vasilatos et al. (2023) and Wang et al. (2023) consider and use perplexity, while Mitchell et al. (2023) use perplexity curvature. Su et al. (2023) use log rank or Tulchinskii et al. (2023) examine the dimensionality of the generated texts. Finally, Yang et al. (2023) use n-grams and analysis based on n-grams.

Further approaches, methods and procedures are presented in the following recent articles (Dhaini et al. 2023; Ghosal et al. 2023; Guo et al. 2023; Tang et al. 2023).

In parallel, the limitations and restrictions of detection approaches are discussed (Helm et al. 2023; Sadasivan et al. 2023; Varshney et al. 2020). They conclude that recognition will become increasingly difficult over time as models become better at imitating human-like speech. By definition, black box recognition will then become impossible. Chakraborty et al. (2023) emphasize that the models can only be recognized by blackbox methods if they are not completely perfect. Accordingly, only enough suitable examples must be available. In summary, the models are still recognizable because they are not yet a perfect representation of human writing. Bhat and Parthasarathy (2020), Wolff and Wolff (2020) and Liyanage and Buscaldi (2023) note that the robustness of detectors may be limited in practice because attackers could use very specific evasion methods (Goodside 2023).

Looking at the existing literature leads to the question of how good the detectors and detector approaches are under real-world conditions (RQ 2). Furthermore, the detectors need to be evaluated on a test dataset that the detectors have never seen before and thus were definitely not part of the training or development of the detector. Several articles deal with the evaluation of detectors and consider AUC or F1 as metrics. In this context, it is not clear whether detectors should be optimized for a low false positive rate or a low false negative rate, which is a trade-off that needs to be discussed based on the use case. In addition, it is important that the detector works equally well on texts from a different of sources and domains. Liang et al. (2023) note that previous test datasets have a simple shape and are strongly related to the training data. Furthermore, Liang et al. (2023) mention that the performance decreases significantly when the test data is from a completely different domain, which indicates that the respective detector is not too robust or cannot be used universally and introduces the need for a relevant and unbiased evaluation dataset for existing methods.

## Research Methodology

Baker (2000) asserts that a meticulously constructed literature review serves as the cornerstone for developing a fresh research endeavor. Additionally, Brocke et al. (2009) highlight that literature reviews have a tendency to uncover previously unknown literature sources, while also playing a crucial role in enhancing the relevance and rigor of research. To document the process of literature search effectively and ensure the attainment of clear and robust outcomes, Brocke et al. (2009) advocate for a five-step framework for literature review. In addition to a comprehensive database search, information systems conferences and journals were also searched for the keywords "AI detection, AI recognition, LLM detection, detection of LLM generated text, GPT detection". The time period was not narrowed down. The result of the literature review can be found in the chapter "Detection of LLMs". After the database search, an intensive search was also conducted for implemented detectors. The following keywords were in the foreground of the database search "GPT detector, ChatGPT detection and LLM detector, AI detector".

To create benchmarking datasets, we consult research on the characteristics of a high-quality dataset. Kistowski et al. (2015) identify relevance, reproducibility, fairness, verifiability, and usability as relevant attributes. Relevance is achieved by evaluating in categories where the need for LLM detection is relevant. Reproducibility, fairness, and verifiability are achieved by performing the same evaluation on the default setting of each detector as an independent research group.

### *Composition of the Evaluation Dataset*

The benchmark dataset for LLM-detection methods should have two main characteristics: relevance to real-world applications and an adequate level of comparability between LLM-generated and non-LLM-generated texts. To ensure relevance, the dataset should focus on areas where deception through LLM-generated texts could cause the greatest harm to society. Therefore, we have created a dataset with product reviews. To create a dataset that allows for comparison between generated and non-generated texts, we aim for homogeneity within a broader category while still allowing for variation. This reduces the chance of duplicate entries but also defines clear boundaries to other categories.

To obtain the product review dataset, we scrape 1000 Amazon phone category reviews written before November 2022. This is done to minimize the possibility of encountering a review written using LLM, because LLM-generation was first widely used after the release of ChatGPT. We then prompt GPT-4 to write 1000 reviews "of a specific phone with 30 to 100 words", ensuring comparability between each entry while allowing for variation. The length of 30 to 100 words matches the typical length of a review and the review length in the Amazon review set. The following review texts show examples of our evaluation dataset.

Example (amazon review): "If your looking for a very stylish and a great smart phone this is the one to get. I got it as a gift and once I got it in my hands and used it for a while I wanted to get one for my self. I recommend getting your hands on one of these amazing smart phones before they are gone."

Example (generated by GPT-4): "I recently purchased the iPhone X and am thoroughly impressed. The display is crisp and vibrant. The Face ID function is seamless and quick. The dual camera system takes stunning photos, even in low light. The only downside is the price which is quite steep."

### *Evaluation Approach*

The texts of the evaluation data set are examined with the two detectors. The texts are transferred individually, and the respective result is noted. Automation via APIs is possible in some cases. For the selected detectors, the answers are percentage values that indicate what percentage of text AI has been generated.

## Results

The following list shows some of the identified detectors: Binoculars, Ghostbuster, GPTZero, Winston AI, Writefull GPT Detector, Writer AI Content Detector, and ZeroGPT. More detectors will be listed in future research. These detectors promise to identify AI generated texts. We choose the detectors based on both their popularity (calculated based on usage statistics) and their academic relevance. Highly specific detectors that do not generalize well are omitted in our analysis. For the benchmarking study, GPTZero and ZeroGPT were selected to start with.

| Metrics | TP | TN | FP | FN | Acc. | Prec. | Rec. | Spec. | F1 |
|---|---|---|---|---|---|---|---|---|---|
| **ZeroGPT** | 732 | 949 | 268 | 51 | 84,05% | 73,20% | 93,49% | 77,98% | 82,11% |
| **GPTZero** | 949 | 1000 | 51 | 0 | 97,45% | 94,90% | 100,00% | 95,15% | 97,38% |

**Table 1. Benchmarking Results for ZeroGPT and GPTZero.**

Table 1 contains the benchmarking results for ZeroGPT and GPTZero. Moreover, the table shows the relevant detector metrics in the context of our evaluation dataset: True Positives (TP), True Negatives (TN), False Positives (FP), False Negatives (FN), Accuracy (Acc.), Precision (Prec.), Recall (Rec.), Specificity (Spec.) and F1-Score.

Table 1 compares the results for the two detectors. It can be clearly seen that the results of GPTZero are better for the developed evaluation data set. The key figures FP and FN in particular are significantly lower than those for the ZeroGPT detector. The 0 cases for false negatives are also remarkable.

Further investigation of these two detectors appears to be appropriate. Among other things, it should be investigated whether the results can also be confirmed for other evaluation data sets. Other evaluation datasets could be, for example, news articles or summaries of scientific articles. In addition, the influence of the length of the text appears to be an interesting influencing factor that should be investigated. The variation in text language should also be considered and analyzed.

## Outlook for Future Research

This contribution gives a good impression of this exciting and current topic. In future research, we focus on providing an overview of LLM detection methods. Future directions could include the following:

- Creation of other relevant benchmarking datasets mentioned (political news, academic writing)
- Multiple LLM detection methods chosen by the factors underlying functionality, popularity, previous results
- Additional LLM's chosen based on performance and popularity
- Analysis and visualization of LLM detection rate, performance of each LLM detector, relationship between sequence length and detectability, susceptibility to adversarial attacks (e.g., several types of paraphrasing), connection between performance and detectability of LLM
- Discussion of the trade-off between false positive and false negative rates
- Discussion of visible trends in the analysis and potential future implications

These aspects are relevant for a comprehensive analysis of the state of AI detection.

## References

Baker, M. J. 2000. "Writing a Literature Review," *The Marketing Review* (1:2), pp. 219-247 (doi: 10.1362/1469347002529189).

Bender, E. M., Gebru, T., McMillan-Major, A., and Shmitchell, S. 2021. "On the Dangers of Stochastic Parrots," in *Proceedings of the 2021 ACM Conference on Fairness, Accountability, and Transparency,* Virtual Event Canada. 03 03 2021 10 03 2021, New York,NY,United States: Association for Computing Machinery, pp. 610-623 (doi: 10.1145/3442188.3445922).

Bhat, M. M., and Parthasarathy, S. 2020. "How Effectively Can Machines Defend Against Machine-Generated Fake News? An Empirical Study," *Proceedings of the First Workshop on Insights from Negative Results in NLP*, pp. 48-53 (doi: 10.18653/v1/2020.insights-1.7).

Biderman, S., and Raff, E. 2022. "Fooling MOSS Detection with Pretrained Language Models," in *Proceedings of the 31st ACM International Conference on Information & Knowledge Management,* New York, NY, USA, New York, NY, USA: ACM (doi: 10.1145/3511808.3557079).

Brocke, J., Alexander Simons, Bjoern Niehaves, Niehaves, B., Reimer, K., Plattfaut, R., and Anne Cleven. 2009. "Reconstucting the giant: on the importance of rigour in documenting the literature search process," *ECIS 2009 Proceedings*.

Chakraborty, S., Bedi, A. S., Zhu, S., an Bang, Manocha, D., and Huang, F. 2023. "On the Possibilities of AI-Generated Text Detection,"

Crothers, E., Japkowicz, N., and Viktor, H. 2022. "Machine Generated Text: A Comprehensive Survey of Threat Models and Detection Methods,"

Dhaini, M., Poelman, W., and Erdogan, E. 2023. "Detecting ChatGPT: A Survey of the State of Detecting ChatGPT-Generated Text,"

Ghosal, S. S., Chakraborty, S., Geiping, J., Huang, F., Manocha, D., and Bedi, A. S. 2023. "Towards Possibilities & Impossibilities of AI-generated Text Detection: A Survey,"

Goodside, R. 2023. "There are adversarial attacks for that proposal as well—in particular, generating with emojis after words and then removing them before submitting defeats it.,"

Guo, B., Zhang, X., Wang, Z., Jiang, M., Nie, J., Ding, Y., Yue, J., and Wu, Y. 2023. "How Close is ChatGPT to Human Experts? Comparison Corpus, Evaluation, and Detection,"

Helm, H., Priebe, C. E., and Yang, W. 2023. "A Statistical Turing Test for Generative Models,"

Hu, X., Chen, P.-Y., and Ho, T.-Y. 2023. "RADAR: Robust AI-Text Detection via Adversarial Learning,"

Kirchenbauer, J., Geiping, J., Wen, Y., Katz, J., Miers, I., and Goldstein, T. 2023. "A Watermark for Large Language Models," *International Conference on Machine Learning*, pp. 17061-17084.

Kistowski, J. von, Arnold, J. A., Huppler, K., Lange, K.-D., Henning, J. L., and Cao, P. 2015. "How to Build a Benchmark," in *Proceedings of the 6th ACMSPEC International Conference on Performance Engineering*, L. K. John (ed.), Austin Texas USA. 28 01 2015 04 02 2015, New York, NY: ACM, pp. 333-336 (doi: 10.1145/2668930.2688819).

Krishna, K., Song, Y., Karpinska, M., Wieting, J., and Iyyer, M. 2023. "Paraphrasing evades detectors of AI-generated text, but retrieval is an effective defense,"

Liang, W., Yuksekgonul, M., Mao, Y., Wu, E., and Zou, J. 2023. "GPT detectors are biased against non-native English writers,"

Liyanage, V., and Buscaldi, D. 2023. "Detecting Artificially Generated Academic Text: The Importance of Mimicking Human Utilization of Large Language Models," in *Natural language processing and information systems*: *28th international conference on applications of natural language to information systems, NLDB 2023, Derby, UK, June 21-23, 2023 : proceedings*, E. Métais, F. Meziane, V. Sugumaran, W. Manning and S. Reiff-Marganiec (eds.), Cham: Springer, pp. 558-565 (doi: 10.1007/978-3-031-35320-8_42).

Mirsky, Y., Demontis, A., Kotak, J., Shankar, R., Gelei, D., Yang, L., Zhang, X., Pintor, M., Lee, W., Elovici, Y., and Biggio, B. 2023. "The Threat of Offensive AI to Organizations," *Computers & Security* (124), p. 103006 (doi: 10.1016/j.cose.2022.103006).

Mitchell, E., Lee, Y., Khazatsky, A., Manning, C. D., and Finn, C. 2023. "DetectGPT: Zero-Shot Machine-Generated Text Detection using Probability Curvature," *International Conference on Machine Learning*, pp. 24950-24962.

Pröhl, T., Mohrhardt, R., Förster, N., Putzier, E., and Zarnekow, R. 2024. "Erkennungsverfahren für KI-generierte Texte: Überblick und Architekturentwurf," *HMD Praxis der Wirtschaftsinformatik*, pp. 1-18 (doi: 10.1365/s40702-024-01051-w).

Pu, J., Huang, Z., Xi, Y., Chen, G., Chen, W., and Zhang, R. 2022. "Unraveling the Mystery of Artifacts in Machine Generated Text," *Proceedings of the Thirteenth Language Resources and Evaluation Conference*, pp. 6889-6898.

Radford, A., Wu, J., Child, R., Luan, D., Amodei, D., and Sutskever, I. 2019. "Language Models are Unsupervised Multitask Learners,"

Sadasivan, V. S., Kumar, A., Balasubramanian, S., Wang, W., and Feizi, S. 2023. "Can AI-Generated Text be Reliably Detected?"

Solaiman, I., Brundage, M., Clark, J., Askell, A., Herbert-Voss, A., Wu, J., Radford, A., Krueger, G., Kim, J. W., Kreps, S., McCain, M., Newhouse, A., Blazakis, J., McGuffie, K., and Wang, J. 2019. "Release Strategies and the Social Impacts of Language Models,"

Su, J., Zhuo, T. Y., Di Wang, and Nakov, P. 2023. "DetectLLM: Leveraging Log Rank Information for Zero-Shot Detection of Machine-Generated Text,"

Tang, R., Chuang, Y.-N., and Hu, X. 2023. "The Science of Detecting LLM-Generated Texts,"

Tian, E. 2023. "New year, new features, new model," *GPTZero*.

Tian, Y., Chen, H., Wang, X., Bai, Z., Zhang, Q., Li, R., Xu, C., and Wang, Y. 2023. "Multiscale Positive-Unlabeled Detection of AI-Generated Texts,"

Tulchinskii, E., Kuznetsov, K., Kushnareva, L., Cherniavskii, D., Barannikov, S., Piontkovskaya, I., Nikolenko, S., and Burnaev, E. 2023. "Intrinsic Dimension Estimation for Robust Detection of AI-Generated Texts,"

Varshney, L. R., Keskar, N. S., and Socher, R. 2020. "Limits of Detecting Text Generated by Large-Scale Language Models,"

Vasilatos, C., Alam, M., Rahwan, T., Zaki, Y., and Maniatakos, M. 2023. "HowkGPT: Investigating the Detection of ChatGPT-generated University Student Homework through Context-Aware Perplexity Analysis,"

Verma, V., Fleisig, E., Tomlin, N., and Klein, D. 2023. "Ghostbuster: Detecting Text Ghostwritten by Large Language Models,"

Wang, Y., Mansurov, J., Ivanov, P., Su, J., Shelmanov, A., Tsvigun, A., Whitehouse, C., Afzal, O. M., Mahmoud, T., Aji, A. F., and Nakov, P. 2023. "M4: Multi-generator, Multi-domain, and Multi-lingual Black-Box Machine-Generated Text Detection,"

Wolff, M., and Wolff, S. 2020. "Attacking Neural Text Detectors,"

Yang, X., Cheng, W., Wu, Y., Petzold, L., Wang, W. Y., and Chen, H. 2023. "DNA-GPT: Divergent N-Gram Analysis for Training-Free Detection of GPT-Generated Text,"

Yu, X., Qi, Y., Chen, K., Chen, G., Yang, X., Zhu, P., Zhang, W., and Yu, N. 2023. "GPT Paternity Test: GPT Generated Text Detection with GPT Genetic Inheritance,"

Zellers, R., Holtzman, A., Rashkin, H., Bisk, Y., Farhadi, A., Roesner, F., and Choi, Y. 2019. "Defending Against Neural Fake News," *Advances in Neural Information Processing Systems* (32).